%% file: main.tex
\documentclass{article}
\usepackage{arxiv}
\usepackage{fontspec}
\usepackage{amsmath,amssymb,booktabs}
\usepackage[numbers,sort]{natbib}
\usepackage[unicode,hidelinks]{hyperref}

\newcommand{\PaperTitle}{Can Semiconductor Virtual Metrology Work with an Off-the-Shelf Model? Evaluating In-Context Learning with TabPFN-3.5}
\renewcommand{\headeright}{}
\renewcommand{\shorttitle}{\footnotesize\PaperTitle}
\renewcommand{\keywordname}{\textbf{Index Terms}}
\renewenvironment{abstract}
  {\begin{center}\bfseries Abstract\end{center}\begin{quote}\noindent\ignorespaces}
  {\end{quote}}

\providecommand{\doi}[1]{doi:~\href{https://doi.org/#1}{\nolinkurl{#1}}}
\hypersetup{
  pdftitle={\PaperTitle},
  pdfauthor={Chi Li, Yu-Ming Hsieh, Fan-Tien Cheng},
  pdflang={en-US},
  pdfkeywords={In-context learning, semiconductor manufacturing, TabPFN, virtual metrology}}

\title{\Large\upshape\bfseries\PaperTitle}
\author{%
  \textbf{Chi Li}\textsuperscript{1}\qquad
  \textbf{Yu-Ming Hsieh}\textsuperscript{2}\qquad
  \textbf{Fan-Tien Cheng}\textsuperscript{1}\\[0.5em]
  \textsuperscript{1}\,Intelligent Manufacturing Research Center National Cheng Kung University (NCKU)\\
  Tainan, Taiwan, R.O.C.\\[0.4em]
  \textsuperscript{2}\,Program on Semiconductor Packaging and Testing\\
  Academy of Innovative Semiconductor and Sustainable Manufacturing\\
  NCKU, Tainan, Taiwan, R.O.C.\\[0.4em]
  {\small
    \href{mailto:lichiisrichard@imrc.ncku.edu.tw}{lichiisrichard@imrc.ncku.edu.tw},\enspace
    \href{mailto:johnniewalk@imrc.ncku.edu.tw}{johnniewalk@imrc.ncku.edu.tw},\enspace
    \href{mailto:chengft@mail.ncku.edu.tw}{chengft@mail.ncku.edu.tw}}}
\date{}

\makeatletter
\renewcommand{\@maketitle}{%
  \vbox{\hsize\textwidth\centering
    \@toptitlebar{\LARGE\@title\par}\@bottomtitlebar
    \vskip 0.6em
    \begin{tabular}{@{}c@{}}\@author\end{tabular}\par}\vskip 1em}
\makeatother

\begin{document}
\maketitle

\begin{abstract}
Developing and maintaining separate virtual metrology models for different semiconductor process conditions often involves training or fine-tuning a model and designing features from sensor traces for each task. Can a single pretrained model predict quality accurately across process conditions without task-specific training or fine-tuning and without designing a feature extraction pipeline for each task? We evaluate TabPFN-3.5 with frozen weights. It uses in-context learning to predict quality from full per-second sensor traces after basic preprocessing, given reference samples with measured quality under the same process condition. TabPFN-3.5 has the lowest mean absolute error of the nine models in each of six conditions spanning three process types. Its error ranges from 34.60 to 68.72 percent of the corresponding standard deviation of measured quality. Replacing full traces with extracted features increases TabPFN-3.5's error in all six conditions but reduces the feedforward neural network's error in four conditions. With full traces, TabPFN-3.5 has lower error on every dataset than any other model evaluated with extracted features. The main drawback is that TabPFN-3.5 takes much longer per prediction than any of the eight baseline models.
\end{abstract}
\begin{quote}\small\noindent
\textbf{\textit{Note to Practitioners}}\enspace Fabs can measure quality for only a fraction of their samples. Each new process condition often means choosing sensor indicators and training and maintaining another model. Engineers can use one off-the-shelf model, TabPFN-3.5, across process conditions. To predict quality for a new sample, engineers give the model its sensor trace and a reference set that pairs sensor traces with measured quality from the same condition. As new measurements arrive, engineers can update the reference set without choosing new indicators or retraining the model. Under each of six production conditions, the model has a smaller average prediction error than all eight comparison methods. For a new process condition, engineers can start with full traces after basic preprocessing. In all six conditions tested, TabPFN-3.5 makes smaller errors with these traces than with compressed summaries of the sensor readings for each recipe step. Adding samples measured under the same condition generally improves accuracy in these tests. The main drawback of TabPFN is that it takes longer to predict quality for each sample. We do not assign lots to folds in production order because poor predictions on later lots could reflect either process drift or the model's failure to learn how sensor readings relate to quality. Our results may therefore overestimate accuracy on later lots. Future work should test whether updating the reference set can sustain accuracy after process drift.
\end{quote}


\section{Introduction}
\label{sec:introduction}

Semiconductor process tools record sensor readings for each sample. Metrology capacity and cost usually limit measurements of quality characteristics such as film thickness and etch depth to a fraction of the samples. Virtual metrology (VM) learns the relationship between sensor traces and measured quality in those samples to estimate the quality of unmeasured samples~\cite{li2025drift},~\cite{hsieh2021cnn}. This relationship can change with production conditions. Li and Hsieh report that when an etching line introduced a new product using the same recipe, a previously weak association between processing time and etch depth became strong, and the existing model's mean absolute error (MAE) rose from 10.03 to 44.34~\cite{li2025drift}. Setting up VM for a different production condition can therefore require engineers to revisit both the features used to represent sensor traces and how they fit the prediction model. We show that a single pretrained model can make accurate VM predictions across six semiconductor production conditions without task-specific training or fine-tuning and without designing a feature extraction pipeline for each task.

Previous studies take different approaches to process traces. Liu et al.\ compress traces with multiway principal component analysis. They then use a long short-term memory (LSTM) network to predict film thickness in chemical vapor deposition~\cite{liu2024hierarchical}. Hsieh et al.\ select sensors and feed their aligned traces into a convolutional neural network (CNN), which learns temporal features for the target process~\cite{hsieh2021cnn}. Using traces directly removes the need to design features by hand. Adapting to a new condition may still require training. For example, the transfer learning method of Hsieh et al.\ requires a few measured samples from a new tool to fine-tune the weights, even when both tools are of the same type~\cite{hsieh2022transfer}. Each task modeled separately gives engineers another model to maintain, and they must keep deciding when it needs retraining.

In-context learning (ICL) offers another way to use measurements. To predict the quality of a new sample, the model uses process traces paired with measured quality from reference samples. TabPFN is a tabular foundation model pretrained on many synthetic tasks. For a new task, TabPFN uses ICL to make predictions from reference samples for that task without updating its weights~\cite{hollmann2023tabpfn},~\cite{hollmann2025foundation}. Two applications of TabPFN closely related to this study are diamond wire slicing of silicon ingots and chemical mechanical polishing (CMP). Zavattari et al.\ treat each cut as one sample. They extract Catch22 features from 20 tool traces and calculate the mean and standard deviation of each trace. They compare the accuracy of several models in predicting the mean and standard deviation of warpage across the silicon slices from each cut. TabPFN gives the most accurate predictions of mean warpage~\cite{zavattari2026warpage}. Chen et al.\ construct 68 features from the mechanics of CMP, including the coupling between pressure in multiple zones and rotational speed, and the cumulative slurry supply. They rank the features using a random forest (RF) and select 20 based on validation performance. With these features fixed, they use only one tenth of the training samples as context. For material removal rate, the test-set coefficient of determination ($R^2$) of TabPFN averages 0.9076 over 20 draws~\cite{chen2026physicscmp}. The first study uses general time series features, and the second uses process-specific features. Each evaluates a single process. This paper asks two questions. First, can TabPFN accurately predict the quality of individual samples directly from per-second sensor readings without feature extraction or selection? Second, can VM tasks under different process conditions share one off-the-shelf model with frozen pretrained weights and achieve accurate per-sample predictions for each task by changing only the reference samples?

We test these questions using TabPFN-3.5 with frozen pretrained weights. For each task, the reference set pairs each measured sample's trace with its measured quality. This set provides the context for predicting the quality of new samples. We use five-fold cross-validation (CV) grouped by lot to compare TabPFN-3.5 with eight baseline models.

The contribution of this paper is to show that TabPFN-3.5 can serve as a common starting VM model across six production conditions, using full sensor traces and reference samples measured under each condition. Engineers can keep the weights and input rules, pair process traces with measured quality under the new condition, and update the reference set as new measurements arrive. We compare prediction accuracy with the full trace and with extracted features. We also examine two questions about deployment. When metrology resources are limited, how do the number and composition of the reference samples affect predictions for unseen lots? How much more inference time does TabPFN-3.5 need than the baseline models to predict quality for the same sample?

\section{Related Work}
\label{sec:related}

\subsection{Model Fitting and Sample Selection in Virtual Metrology}
VM studies typically use measured samples to fit a model for a specific quality target. Choi and Hong use a feedforward neural network (FNN) to predict film thickness in amorphous carbon deposition~\cite{choi2021film}. Deivendran et al.\ combine a physics model with XGBoost to estimate the material removal rate in CMP~\cite{deivendran2025cmp}. Chien et al.\ use an isolation forest to group data from multiple tools and recipes, then fit a separate RF regression model for each group~\cite{chien2022decision}. Ji et al.\ use a Transformer to predict process quality~\cite{ji2025alignment}. Our comparison covers these model types, the CNN and LSTM described in the introduction, and two models adapted from other virtual sensing studies. The latter are a temporal convolutional network (TCN) and a graph neural network (GNN)~\cite{gumbarevic2026heat},~\cite{zhao2025virtual}.

Which samples are measured and which measured samples are kept for model updates can both affect the relationship a model learns. Hsieh et al.\ retain samples at both ends of the quality distribution to preserve the model's ability to predict rare extremes~\cite{hsieh2024samples}. When process conditions change, Lin et al.\ use prediction uncertainty to select samples for measurement, then fine-tune a pretrained model with the new measurements~\cite{lin2025fewshot}. These studies consider how to update a model with limited metrology resources. In ICL with frozen pretrained weights, the choice of measured samples directly determines which process traces and measured quality values the model can use at each prediction.

\subsection{From Prior-Data Fitted Networks to Tabular Regression}
TabPFN grew out of the prior-data fitted network (PFN) proposed by M\"uller et al. The researchers repeatedly generate supervised learning tasks from a specified prior. For each task, they provide some samples with their labels and train the network to predict the labels of the remaining samples. Through pretraining, the network learns to approximate the posterior predictive distribution under that prior given the observed samples. Observations from a new task change the predictions while the weights remain fixed~\cite{muller2022bayesian}.

Hollmann et al.\ released the first TabPFN in 2022 and presented it at the International Conference on Learning Representations (ICLR 2023). They generate tabular classification tasks from a prior that includes structural causal models. Their evaluation focuses on tasks with up to 1,000 training samples, 100 numerical columns, and 10 classes~\cite{hollmann2023tabpfn}. The second generation extends ICL to regression with continuous targets and was evaluated on up to 10,000 samples and 500 columns~\cite{hollmann2025foundation}. With regression, continuous measurements such as film thickness or etch depth can serve directly as targets. In other engineering applications, researchers use TabPFN to predict tensile strength retention in fiber-reinforced composites at elevated temperatures~\cite{nikzad2026frp} and the non-steady-state chloride migration coefficient of 3D-printed concrete~\cite{liu2026concrete}.

\subsection{Number of Reference Samples and Table Width}
When process traces are represented as a table, the number of rows determines how many measured samples the model can use as references. The number of columns determines how many time and sensor positions the table can retain per sample. TabPFN-2.5 is designed for up to 50,000 samples and 2,000 columns~\cite{grinsztajn2025tabpfn25}. TabPFN-3 improves attention and context caching and was evaluated on tables with up to one million samples. When the same reference data are used to predict multiple samples, the cache lets the model reuse some computations from building the context~\cite{grinsztajn2026tabpfn3}.

TabPFN-3.5 expands the range of tasks with wide tables and includes distribution differences between groups in pretraining and evaluation. It uses the same pretrained weights for classification and regression. The recommended input width increased from 100 columns in v1 and 500 in v2 to 6,000 in v3.5. With more estimators, v3.5 supports up to 20,000 columns~\cite{jager2026tabpfn35}. Support for wide tables lets us use per-second sensor readings as input. We use the same model to compare prediction accuracy with full traces and with features extracted by applying principal component analysis to step-level sensor statistics.

\input{method-results}

\section{Conclusions and Future Work}
\label{sec:conclusion}

This paper asks whether VM tasks under different process conditions can share one off-the-shelf model without feature extraction or per-task training. Our results on six production tasks support this. TabPFN-3.5 keeps the same pretrained weights on all six tasks. For each task, it receives only the preprocessed full trace and the reference samples measured under that condition. In five-fold CV grouped by lot, it has the lowest MAE of the nine models on every task, from 34.60\% to 68.72\% of the standard deviation of measured quality ($\sigma_Y$). We select the baseline hyperparameters on the same five folds but use one fixed setting for TabPFN-3.5, so any selection bias from this step favors the baseline models only. These results support using TabPFN-3.5 as a common starting VM model for each new process condition.

Replacing the full trace with principal components of step-level statistics raises MAE for TabPFN-3.5 on all six tasks. For FNN, the same lower-dimensional input reduces MAE on four tasks. On these tasks, the per-second readings may hold information about quality that step-level statistics and their principal components do not keep. These findings apply only to the feature extraction pipeline evaluated here. Whether features designed for a specific quality target or fine-tuning TabPFN-3.5 on production data can improve accuracy remains an open question.

With the weights fixed, the reference set is what engineers maintain for each task. Both its size and its composition affect accuracy. On a fixed test set of held-out lots, the full reference pool gives lower MAE than the median MAE from 30 random draws of 30 reference samples on all six tasks. In these experiments, increasing the number of reference samples is the most consistent way to reduce MAE. With 30 reference samples, the difference between the highest and lowest MAE across the draws ranges from 11.60\% to 33.45\% of $\sigma_Y$. Under a fixed metrology budget, deciding which samples to measure therefore becomes part of building the VM model. Future work could test whether selecting samples by trace similarity to the new sample, or by the process conditions they cover, gives lower MAE than random draws of the same size.

The main cost is inference time. TabPFN-3.5 takes 756.75 to 5,421.56 ms to predict one sample, whereas each baseline model takes at most 91.93 ms. When several tasks share one GPU, transferring a task's cache from host memory adds 110.92 to 766.05 ms per sample. Each update of the reference set also requires rebuilding the cache, which takes 2,037.62 to 24,365.11 ms. Sharing one model across tasks thus trades longer per-sample inference for skipping per-task training and feature design. Our use of the pretrained TabPFN-3.5 weights under Prior Labs' non-commercial license is limited to this offline feasibility study.

We do not order the folds by production time. We use this split to limit the influence of process drift when comparing how well the models learn the relationship between sensor readings and quality. The reference set may therefore include samples from lots produced after the test lots, and CV may overestimate accuracy on later lots. How prediction error varies across lots evaluated in production order and when to update the reference set are both open questions.

\section*{Acknowledgment}
The authors used Claude Opus 5.5 (Anthropic) and GPT-6.2 (OpenAI) to translate the manuscript into English and to edit the translation for clarity and style.

\input{main.bbl}

\end{document}

%% file: method-results.tex
\section{Methods}
\label{sec:methods}

Each reference sample contains the sensor trace from one wafer during one process operation and its measured quality. We first arrange the per-second readings into traces of a fixed length. We then use reference samples from the same process condition as context to predict the quality of a new sample.

\subsection{From Process Traces to Per-Second Inputs}
\label{sec:trace-input}

We collect sensor readings for each sample at one-second intervals during processing. We exclude traces with outlying processing times using the interquartile range (IQR) of processing times in each dataset and the $3\times\mathrm{IQR}$ rule. We then process each trace in four steps.

\begin{enumerate}
  \item \textbf{Set the common time window and truncate long traces.} We set the window length to the 99th percentile of processing time in seconds for the reference samples. We truncate each trace to retain only sensor readings within this window.
  \item \textbf{Impute missing values during processing.} For each sensor, we pool all valid readings from the reference samples within the common time window and calculate their median. We use that sensor's median to impute missing values during processing in the traces of both reference samples and new samples.
  \item \textbf{Standardize the sensor readings.} For each sensor, we calculate the mean and standard deviation from the imputed reference data. We subtract this mean from each reading and divide by this standard deviation for both reference samples and new samples.
  \item \textbf{Zero-pad short traces.} If processing ends early, we zero-pad the standardized trace at the end to fill the common time window. We exclude these zeros when estimating the median, mean, and standard deviation.
\end{enumerate}

We preprocess each new sample using the same settings as for the reference samples. In the small-sample experiment (Section~\ref{sec:reference-size}), we estimate the time window and statistics using only the reference samples selected in each draw.

\subsection{In-Context Regression with Paired Reference Samples}

TabPFN-3.5 uses multiple estimators, each of which runs inference on a different permutation of the same dataset~\cite{jager2026tabpfn35}. All estimators in this study share the same pretrained weights and reference samples. Each estimator uses its assigned columns to estimate the quality distribution for the same new sample. The model then combines these distributions into one predictive distribution.

For a given reference set, let $C$ be the number of sensors and $T$ the number of time steps in the common time window. We flatten each trace in time order, using a fixed sensor order within each second. Each row represents one sample, and each column corresponds to one sensor at one time step. This representation requires no feature extraction or selection. The total number of columns after flattening, $P$, is
\begin{equation}
  \label{eq:input-width}
  P=C\times T .
\end{equation}

We assign each estimator at most 768 of the columns. Let $M$ be the number of estimators and $n$ the number of columns each estimator receives. We use at least 8 estimators and add more as the number of input columns grows so that the ensemble covers the full trace. In the following expressions, $\lceil P/768\rceil$ rounds $P/768$ up to the nearest integer. The functions $\max$ and $\min$ return the larger and smaller of the two arguments, respectively.
\begin{align}
  M &= \max\!\left\{8,\left\lceil P/768\right\rceil\right\}, \label{eq:members}\\
  n &= \min\{P,768\}. \label{eq:member-width}
\end{align}
When $P$ is at most 768, each estimator receives all columns. When $P$ exceeds 768, the model randomly orders the columns and assigns them to the estimators in sequence. Only after assigning all $P$ columns once does it draw from previously assigned columns to fill each estimator's allocation. More than one estimator can therefore receive the same column. This rule assigns every input column to at least one estimator. We set \texttt{ignore\_pretraining\_limits=True} to retain the full trace when the number of columns exceeds the recommended size for the pretrained model. We disable the fingerprint feature enabled by default and keep polynomial features disabled. The input thus contains only the preprocessed per-second sensor readings.

During pretraining, TabPFN draws synthetic tasks from a prior over data-generating mechanisms. It learns to approximate the posterior predictive distribution under that prior given the observed samples~\cite{hollmann2023tabpfn},~\cite{muller2022bayesian}. Each VM task provides a reference set $D$. Let $i$ index the samples. Each pair $(\mathbf{x}_i,y_i)$ in $D$ contains the flattened input vector $\mathbf{x}_i$ and measured quality $y_i$ for sample $i$. For a given test sample, let $\mathbf{x}_*$ be its flattened input vector and $y_*$ its quality, which has yet to be measured. Let $p(y_*\mid\mathbf{x}_*,D)$ denote the posterior predictive density of its quality under the prior, given the trace $\mathbf{x}_*$ and reference set $D$. TabPFN uses the same pretrained weights $\theta$ for all six tasks and combines the estimators' predictions to approximate this density as $p_\theta(y_*\mid\mathbf{x}_*,D)$. The model uses the conditional mean of this distribution as its point estimate of $y_*$, denoted by $\widehat{y}_*$~\cite{jager2026tabpfn35}. When the process condition changes, we replace $D$ with the reference set for that task and keep $\theta$ fixed.

We fix the random seed and use the \texttt{tabpfn-v3.5-20260909.safetensors} checkpoint. To build the key--value (KV) cache for reference set $D$, we use \texttt{fit\_mode="fit\_with\_cache"}. The cache stores both keys and values in an 8-bit floating-point format (FP8 E4M3, \texttt{torch.float8\_e4m3fn}). We can reuse this cache for subsequent samples and rebuild it when the reference set changes.

\section{Results and Discussion}
\label{sec:results}

\subsection{Production Datasets}

The production data cover six process conditions across three types of processes. Each condition defines one VM task. Diffusion (DIFF) uses heat treatment to make atoms diffuse within a material. We denote the four diffusion conditions by DIFF\textsubscript{A} through DIFF\textsubscript{D}. Etching (ETCH) selectively removes material from the substrate surface to form the desired pattern. The thin film (TF) process forms the material layers that devices require on the substrate surface.

The six datasets contain 3,622 samples in total, each pairing a wafer's sensor trace from the operation covered by its dataset with the corresponding quality measurement. Table~\ref{tab:datasets} reports the number of paired samples and the full-trace input dimensions for each dataset.

\begin{table}[htbp]
  \centering
  \small
  \setlength{\tabcolsep}{3pt}
  \caption{Paired samples and full-trace input dimensions for the six production datasets}
  \label{tab:datasets}
  \begin{tabular}{@{}lrrrr@{}}
    \toprule
    Dataset & Paired samples & Sensors ($C$) & Time steps ($T$) & Flattened columns ($P$) \\
    \midrule
    DIFF\textsubscript{A} & 495 & 57 & 230 & 13,110 \\
    DIFF\textsubscript{B} & 274 & 57 & 231 & 13,167 \\
    DIFF\textsubscript{C} & 696 & 57 & 244 & 13,908 \\
    DIFF\textsubscript{D} & 241 & 57 & 242 & 13,794 \\
    ETCH & 1,226 & 53 & 441 & 23,373 \\
    TF & 690 & 26 & 161 & 4,186 \\
    \bottomrule
  \end{tabular}
\end{table}

\subsection{Computing Environment}

We run all experiments on one Windows 11 computer with an Intel Core i7-12700F CPU and an NVIDIA GeForce RTX 3080 GPU. TabPFN-3.5 builds the context and runs inference on the GPU using Python 3.12.12, TabPFN package version 10.2.0, and PyTorch 2.9.1 (CUDA 12.8). The six neural network baselines run on the same GPU using PyTorch 2.4.0. RF and XGBoost run on the CPU using scikit-learn 1.8.0 and XGBoost 3.1.3, respectively.

\subsection{Prediction Accuracy of TabPFN-3.5 and Eight Baselines Using Full Traces}
\label{sec:model-comparison}

We compare nine models using the same full traces after the preprocessing in Section~\ref{sec:trace-input}. For TabPFN-3.5, FNN, RF, and XGBoost, we flatten each trace into one row of a table. CNN, TCN, and LSTM take each trace as a time-by-sensor array. The Transformer linearly projects the readings in each nonoverlapping four-second window into one token. The GNN treats each sensor as a node and encodes its time series with a gated recurrent unit (GRU).

For each task, we use five-fold CV with a fixed random seed and keep all samples from each lot in the same fold to prevent leakage. We form the folds without ordering lots by production time to compare models on held-out lots across the observed production period. In a chronological split, a high MAE on later lots could arise because the model struggled to learn how sensor readings relate to quality or because process drift changed a relationship the model had learned well on earlier lots.

All nine models use the same folds. In each round, we use four folds to fit the baseline models or build the reference set for TabPFN-3.5. We evaluate all models on the same held-out fold. For each round, we estimate the common time window from the four training folds. In ETCH, the longest training window spans 444 time steps, yielding 23,532 flattened columns per sample.

For each task, we combine each model's predictions from the five held-out folds and calculate MAE over all samples. Table~\ref{tab:wafer} lists the results. We select the baseline hyperparameters using MAE on these same five folds, so selection bias may affect the reported baseline errors. TabPFN-3.5 has lower MAE than all eight baseline models on each of the six tasks. Among the eight models other than TabPFN-3.5, XGBoost has the lowest MAE on DIFF\textsubscript{A}, DIFF\textsubscript{B}, DIFF\textsubscript{C}, and ETCH, and RF has the lowest MAE on DIFF\textsubscript{D} and TF. In the table, $\sigma_Y$ denotes the standard deviation of all measured quality values in each dataset.

\begin{table}[htbp]
  \centering
  \small
  \setlength{\tabcolsep}{1.8pt}
  \caption{MAE in five-fold cross-validation on six production tasks}
  \label{tab:wafer}
  \begin{tabular}{@{}lrrrrrrrrrr@{}}
    \toprule
    Dataset & $\sigma_Y$ & TabPFN & CNN & LSTM & TCN & Transformer & GNN & FNN & RF & XGBoost \\
    \midrule
    DIFF\textsubscript{A} & 1.095 & \textbf{0.4017} & 0.7373 & 0.7857 & 0.7045 & 0.7350 & 0.8630 & 0.7583 & 0.6033 & 0.5722 \\
    DIFF\textsubscript{B} & 1.076 & \textbf{0.3722} & 0.6700 & 0.7182 & 0.5989 & 0.6046 & 0.7748 & 0.6504 & 0.4930 & 0.4632 \\
    DIFF\textsubscript{C} & 0.3067 & \textbf{0.1667} & 0.2189 & 0.2273 & 0.2037 & 0.2126 & 0.2287 & 0.2049 & 0.1906 & 0.1875 \\
    DIFF\textsubscript{D} & 0.2504 & \textbf{0.1721} & 0.2019 & 0.1988 & 0.1976 & 0.1959 & 0.1985 & 0.2002 & 0.1846 & 0.1858 \\
    ETCH & 20.84 & \textbf{13.36} & 15.12 & 15.77 & 15.18 & 15.24 & 15.52 & 14.04 & 14.13 & 13.39 \\
    TF & 29.03 & \textbf{19.73} & 20.97 & 21.50 & 21.52 & 21.41 & 21.49 & 21.33 & 20.43 & 20.60 \\
    \bottomrule
  \end{tabular}
  \par\smallskip
\end{table}

Across the six tasks, the MAE of TabPFN-3.5 ranges from 34.60\% to 68.72\% of the standard deviation of measured quality ($\sigma_Y$) in each dataset. DIFF\textsubscript{B} has the lowest percentage and DIFF\textsubscript{D} the highest. These results show that TabPFN-3.5 can provide useful predictions directly from per-second sensor readings after basic preprocessing, with its pretrained weights fixed.

\subsection{Context Cache Build, Transfer, and Per-Sample Inference Times}
\label{sec:timing}

Fabs often support several process conditions at once. It is difficult to fit the context caches for all tasks in the memory of a single GPU. In the following experiments, the default is to keep each task's FP8 KV cache in host memory and transfer the required cache to the GPU only when predicting quality for a sample.

The timing experiment uses the lot groups from Section~\ref{sec:model-comparison}. We hold out fold 1, which contains about 20\% of the lots in each task. We build the cache using the full reference pool from the other four folds and time per-sample inference on the held-out fold. Table~\ref{tab:latency} lists the times for three stages (cache building, cache transfer, and remaining per-sample inference). The first stage covers processing the reference data with the weights already loaded and packing the cache on the CPU. The second measures the time to transfer the cache from the CPU to the GPU for each sample. The third measures the remaining per-sample inference time, starting with a standardized input for one sample. It includes internal preprocessing, the forward pass, combining the estimators' predictions, and returning the result. The timings exclude weight transfers and the transfer of the newly built cache back to the CPU.

\begin{table}[htbp]
  \centering
  \small
  \caption{Sample counts and times for the three stages of TabPFN-3.5 on six tasks}
  \label{tab:latency}
  \begin{tabular}{@{}lrrrrr@{}}
    \toprule
    Dataset & All paired & Reference & Build cache & Transfer cache & Inference \\
            & samples & samples & (ms/update) & (ms/sample) & (ms/sample) \\
    \midrule
    DIFF\textsubscript{A} & 495 & 393 & 4,875.52 & 447.78 & 2,240.70 \\
    DIFF\textsubscript{B} & 274 & 222 & 4,360.85 & 383.06 & 2,148.77 \\
    DIFF\textsubscript{C} & 696 & 554 & 6,162.75 & 389.06 & 2,176.79 \\
    DIFF\textsubscript{D} & 241 & 191 & 3,928.07 & 385.17 & 2,297.32 \\
    ETCH & 1,226 & 985 & 24,365.11 & 766.05 & 5,421.56 \\
    TF & 690 & 552 & 2,037.62 & 110.92 & 756.75 \\
    \bottomrule
  \end{tabular}
\end{table}

For all six tasks in Table~\ref{tab:latency}, building the cache takes longer per update than inference takes per sample. Per-sample inference in turn takes longer than transferring the cache per sample. Table~\ref{tab:warm-latency} compares per-sample inference times for the nine models on the same samples. RF and XGBoost run on the CPU, and the other models run on the GPU. TabPFN-3.5 has longer per-sample inference times than all eight baseline models on each of the six tasks. This additional latency recurs with every prediction, making it the main practical drawback of the ICL-based VM approach evaluated here.

\newpage
\begin{table}[htbp]
  \centering
  \small
  \setlength{\tabcolsep}{3pt}
  \caption{Per-sample inference time for the nine models (ms/sample)}
  \label{tab:warm-latency}
  \begin{tabular}{@{}lrrrrrrrrr@{}}
    \toprule
    Dataset & TabPFN & CNN & LSTM & TCN & Transformer & GNN & FNN & RF & XGBoost \\
    \midrule
    DIFF\textsubscript{A} & 2,240.70 & 4.33 & 6.24 & 0.94 & 1.25 & 4.14 & 0.55 & 46.65 & 0.34 \\
    DIFF\textsubscript{B} & 2,148.77 & 3.62 & 5.90 & 0.79 & 1.24 & 3.68 & 0.38 & 46.80 & 0.37 \\
    DIFF\textsubscript{C} & 2,176.79 & 3.37 & 5.39 & 0.93 & 1.25 & 4.54 & 0.38 & 46.63 & 0.40 \\
    DIFF\textsubscript{D} & 2,297.32 & 3.49 & 5.63 & 0.81 & 1.31 & 4.33 & 0.30 & 89.65 & 0.39 \\
    ETCH & 5,421.56 & 7.23 & 11.13 & 0.90 & 1.59 & 7.92 & 0.52 & 91.93 & 0.45 \\
    TF & 756.75 & 1.70 & 5.30 & 1.01 & 2.48 & 3.80 & 0.57 & 89.27 & 0.30 \\
    \bottomrule
  \end{tabular}
\end{table}

\subsection{Comparison of the Full Trace and Extracted Features}
\label{sec:input-comparison}

How does each model's error change when we replace full traces with extracted features? The full trace preserves the position of each reading by second and sensor. To obtain extracted features, we convert the same trace into step-level statistics and reduce their dimension using principal component analysis (PCA). We choose TabPFN-3.5, FNN, RF, and XGBoost because all four accept both the flattened full trace and the extracted features as tabular input. The other five models are better suited to inputs that preserve temporal order or sensor-graph structure. The principal components have no direct correspondence to time steps or sensor nodes. We use the same five folds of paired samples as in Section~\ref{sec:model-comparison}. In each fold, both representations use the same training samples, and we compare MAE on the same held-out samples.

We extract features from the per-second sensor readings described in Section~\ref{sec:trace-input}. For each sample, we group the readings by recipe step. For each sensor within each step, we calculate the maximum, mean, minimum, range, slope over time, and standard deviation. We also record the number of time steps in each recipe step. For example, DIFF\textsubscript{A} has 13 recipe steps and 57 sensors, yielding $13\times(57\times6+1)=4{,}459$ candidate statistic columns per sample.

To fit a PCA model for each task, we use samples that have not been selected for measurement under the same process condition. We call these unpaired samples because their sensor traces have no corresponding measured quality. To select the number of principal components, we split the unpaired samples into five folds grouped by lot. In each round, we estimate imputation and standardization parameters using only the four training folds. We remove constant columns and fit PCA to the standardized step-level statistics. We select the number of principal components that gives the lowest reconstruction error on the held-out fold. We average the five selected counts and round the result to the nearest integer. Using that count, we re-estimate the preprocessing parameters and fit the final PCA model using all unpaired samples. We then project the paired samples for that task onto the same principal component space.

\begin{table}[htbp]
  \centering
  \caption{Input dimensions for the full trace and extracted features}
  \label{tab:stepdims}
  \small
  \setlength{\tabcolsep}{3.4pt}
  \begin{tabular}{@{}lrrrrr@{}}
    \toprule
    Dataset & \shortstack[r]{Flattened\\columns ($P$)} & \shortstack[r]{Recipe\\steps} & \shortstack[r]{Candidate\\statistic columns} & \shortstack[r]{PCA input\\columns} & \shortstack[r]{Final principal\\components} \\
    \midrule
    DIFF\textsubscript{A} & 13,110 & 13 & 4,459 & 4,022 & 2,672 \\
    DIFF\textsubscript{B} & 13,167 & 13 & 4,459 & 4,027 & 2,639 \\
    DIFF\textsubscript{C} & 13,908 & 13 & 4,459 & 3,846 & 2,816 \\
    DIFF\textsubscript{D} & 13,794 & 13 & 4,459 & 3,932 & 2,054 \\
    ETCH & 23,373 & 15 & 4,785 & 4,769 & 3,965 \\
    TF & 4,186 & 15 & 2,355 & 1,475 & 987 \\
    \bottomrule
  \end{tabular}
\end{table}

Table~\ref{tab:stepdims} shows how the input dimensions change after feature extraction. Across the six tasks, the relative decrease in column count ranges from 76.42\% to 85.11\%. For ETCH, the full trace has 23,373 columns of per-second readings. Computing step-level statistics yields 4,785 candidate columns, of which 4,769 enter PCA. We retain 3,965 principal components. For TF, the input decreases from 4,186 columns to 987 principal components.

Table~\ref{tab:stepmetrics} compares MAE for the two input representations within each model and task. TabPFN-3.5 uses the same frozen pretrained weights for both representations. We fit FNN, RF, and XGBoost separately for each representation. In the table, Full trace denotes the complete per-second readings, and Extracted features denotes the principal components of the step-level statistics.

\begin{table}[htbp]
  \centering
  \caption{MAE of four models using the full trace or extracted features}
  \label{tab:stepmetrics}
  \small
  \begin{tabular}{@{}llrrrrrr@{}}
    \toprule
    Model & Input representation & DIFF\textsubscript{A} & DIFF\textsubscript{B} & DIFF\textsubscript{C} & DIFF\textsubscript{D} & ETCH & TF \\
    \midrule
    TabPFN-3.5 & Full trace & \textbf{0.4017} & \textbf{0.3722} & \textbf{0.1667} & \textbf{0.1721} & \textbf{13.36} & \textbf{19.73} \\
     & Extracted features & 0.7273 & 0.6698 & 0.2007 & 0.1903 & 15.33 & 20.73 \\
    \addlinespace[1.5pt]
    FNN & Full trace & 0.7583 & 0.6504 & 0.2049 & 0.2002 & \textbf{14.04} & \textbf{21.33} \\
     & Extracted features & \textbf{0.7057} & \textbf{0.5789} & \textbf{0.1963} & \textbf{0.1922} & 15.84 & 21.41 \\
    \addlinespace[1.5pt]
    RF & Full trace & \textbf{0.6033} & \textbf{0.4930} & \textbf{0.1906} & \textbf{0.1846} & \textbf{14.13} & \textbf{20.43} \\
     & Extracted features & 0.7566 & 0.7182 & 0.2177 & 0.1938 & 15.29 & 20.88 \\
    \addlinespace[1.5pt]
    XGBoost & Full trace & \textbf{0.5722} & \textbf{0.4632} & \textbf{0.1875} & \textbf{0.1858} & \textbf{13.39} & \textbf{20.60} \\
     & Extracted features & 0.7406 & 0.7004 & 0.2140 & 0.1980 & 15.36 & 20.79 \\
    \bottomrule
  \end{tabular}
  \par\smallskip
\end{table}

With extracted features, the MAE of TabPFN-3.5 increases on all six tasks. The relative increase ranges from 5.072\% on TF to 81.07\% on DIFF\textsubscript{A}. On DIFF\textsubscript{B}, the same feature extraction increases the MAE of TabPFN-3.5 from 0.3722 to 0.6698 but reduces FNN's MAE from 0.6504 to 0.5789. FNN's MAE falls on DIFF tasks, with relative decreases ranging from 3.96\% to 10.99\%. Feature extraction increases the MAE of both RF and XGBoost on all six tasks. On each of the six tasks, TabPFN-3.5 with the full trace also has lower MAE than the best model using extracted features.

These conclusions apply to the specific feature extraction pipeline we evaluated. The pipeline replaces the per-second readings with step-level statistics, estimates imputation and standardization parameters from unpaired samples, and projects the standardized statistics onto principal components. We set the number of components based on reconstruction error. For TabPFN-3.5, the smaller inputs produced by this basic pipeline came at the cost of prediction accuracy. We cannot rule out the possibility of accuracy gains from features designed for a specific quality target.

\subsection{How the Number and Composition of Reference Samples Affect Prediction Accuracy}
\label{sec:reference-size}

Fabs have limited metrology capacity, so each measured sample matters. Measuring another sample can provide one more reference sample for TabPFN-3.5, but it adds to metrology costs. On each of the six tasks, we fix the model weights and the test samples. We examine how prediction error changes as the number of reference samples increases. We also make repeated draws at the same size to measure how much MAE varies with the choice of samples.

For each task, we use the lot grouping from Section~\ref{sec:model-comparison}. We hold out fold 1 (about 20\% of the lots) as a fixed test set and use the other four folds as the reference pool. We set ten context sizes from 30 samples to the full reference pool using a geometric sequence rounded to integers. We evaluate the full reference pool once and make 30 random draws at each other size. For each draw, we use only the selected samples to estimate the time window (the 99th percentile of processing time), imputation medians, and standardization parameters.

\begin{figure}[!htb]
  \centering
  \leavevmode\XeTeXpdffile "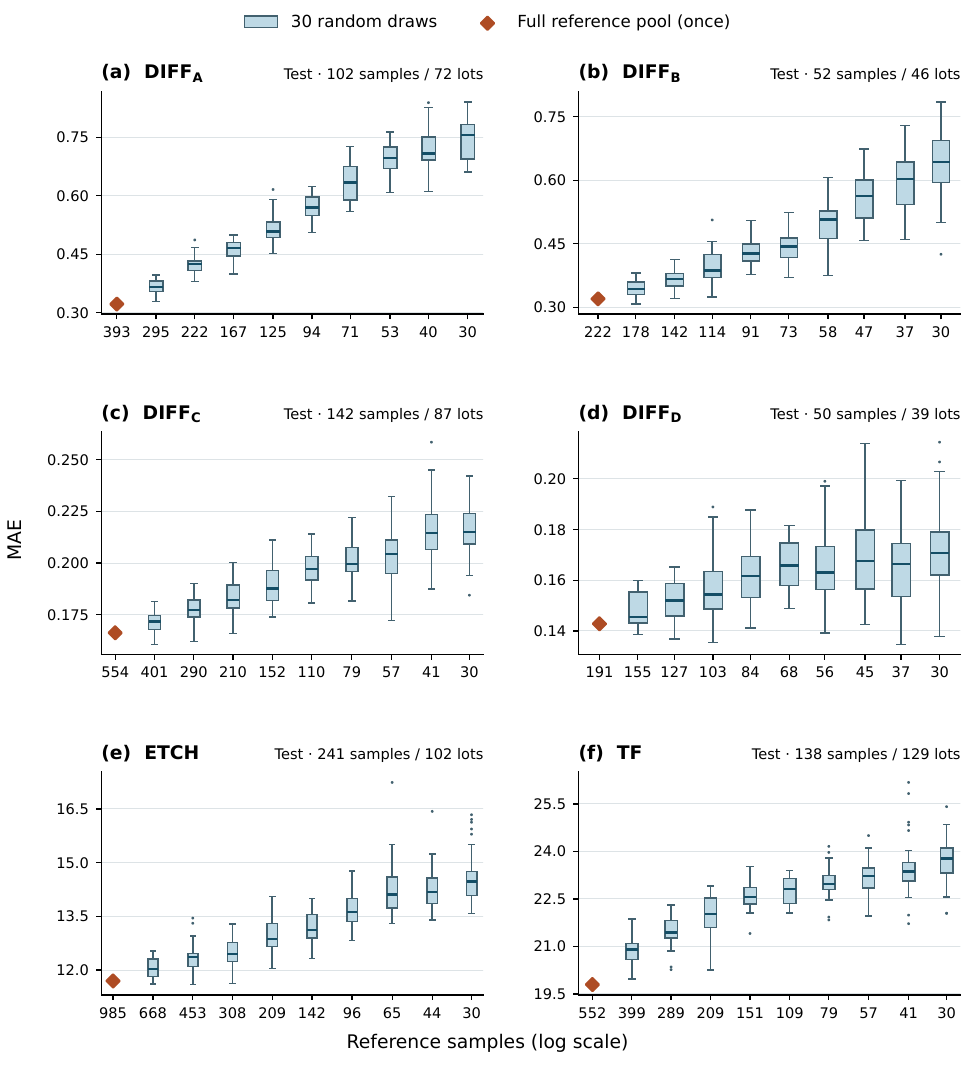" width 0.83\linewidth\relax
  \caption{MAE by number of reference samples on the six datasets}
  \label{fig:context-sizes}
\end{figure}

In Fig.~\ref{fig:context-sizes}, the number of reference samples decreases from left to right along the logarithmic horizontal axis. The vertical axis shows MAE. The labels in the upper right of each panel give the sample and lot counts in the fixed test set. The leftmost diamond shows the single result for the full reference pool. Each box plot shows the distribution of MAE across 30 random draws. The box spans the 25th to 75th percentiles, and the line inside marks the median. Each whisker extends to the most distant observation no more than 1.5 interquartile ranges beyond the corresponding box edge. Observations beyond the whiskers appear as individual points.

The box plots also show that MAE varies with the choice of samples even when the number of reference samples is the same. For DIFF\textsubscript{B} in Fig.~\ref{fig:context-sizes}(b), we draw 37 samples each time and predict quality for the same 52 test samples. Across 30 draws, the 25th and 75th percentiles of MAE are 0.5426 and 0.6427, respectively. This variation includes the effect of re-estimating preprocessing parameters from the samples selected in each draw.

Table~\ref{tab:context-size} shows how MAE varies across 30 draws of 30 reference samples on the same test set. For DIFF\textsubscript{B}, MAE ranges from 0.4251 to 0.7850, a difference of 0.3599 or about one-third of $\sigma_Y$. Across tasks, the difference between maximum and minimum MAE ranges from 11.60\% to 33.45\% of $\sigma_Y$. The full reference pool yields a lower MAE than the median from 30-sample draws on every task. Whether selecting references by trace similarity or coverage of process conditions consistently outperforms random draws under a fixed metrology budget remains open.

\begin{table}[!htb]
  \centering
  \caption{Variation in MAE across repeated draws of 30 reference samples and MAE with the full reference pool}
  \label{tab:context-size}
  \small
  \setlength{\tabcolsep}{4pt}
  \begin{tabular}{@{}lrrrrrrr@{}}
    \toprule
    & \multicolumn{2}{c}{Samples} & & \multicolumn{3}{c}{\shortstack{MAE with 30 reference samples\\(30 draws)}} & Full pool \\
    \cmidrule(lr){2-3}\cmidrule(lr){5-7}\cmidrule(l){8-8}
    Dataset & Fixed test set & Reference pool & $\sigma_Y$ & Median & Min.--max. & Range/$\sigma_Y$ & MAE \\
    \midrule
    DIFF\textsubscript{A} & 102 & 393 & 1.095 & 0.7556 & 0.6607--0.8412 & 16.48\% & 0.3222 \\
    DIFF\textsubscript{B} & 52 & 222 & 1.076 & 0.6424 & 0.4251--0.7850 & 33.45\% & 0.3198 \\
    DIFF\textsubscript{C} & 142 & 554 & 0.3067 & 0.2152 & 0.1845--0.2420 & 18.77\% & 0.1663 \\
    DIFF\textsubscript{D} & 50 & 191 & 0.2504 & 0.1708 & 0.1378--0.2145 & 30.61\% & 0.1428 \\
    ETCH & 241 & 985 & 20.84 & 14.48 & 13.59--16.34 & 13.21\% & 11.70 \\
    TF & 138 & 552 & 29.03 & 23.78 & 22.05--25.41 & 11.60\% & 19.80 \\
    \bottomrule
  \end{tabular}
  \par\smallskip
\end{table}

%% file: main.bbl

%% file: main.bbl
\begin{thebibliography}{10}
\providecommand{\url}[1]{#1}
\csname url@samestyle\endcsname
\providecommand{\newblock}{\relax}
\providecommand{\bibinfo}[2]{#2}
\providecommand{\BIBentrySTDinterwordspacing}{\spaceskip=0pt\relax}
\providecommand{\BIBentryALTinterwordstretchfactor}{4}
\providecommand{\BIBentryALTinterwordspacing}{\spaceskip=\fontdimen2\font plus
\BIBentryALTinterwordstretchfactor\fontdimen3\font minus
  \fontdimen4\font\relax}
\providecommand{\BIBforeignlanguage}[2]{{%
\expandafter\ifx\csname l@#1\endcsname\relax
\typeout{** WARNING: IEEEtran.bst: No hyphenation pattern has been}%
\typeout{** loaded for the language `#1'. Using the pattern for}%
\typeout{** the default language instead.}%
\else
\language=\csname l@#1\endcsname
\fi
#2}}
\providecommand{\BIBdecl}{\relax}
\BIBdecl

\bibitem{li2025drift}
C.~Li and Y.-M. Hsieh, ``Sudden concept drift detection and adaptation in
  virtual metrology for semiconductor manufacturing,'' \emph{IEEE Access},
  vol.~13, pp.~151\,614--151\,623, 2025.

\bibitem{hsieh2021cnn}
Y.-M. Hsieh, T.-J. Wang, C.-Y. Lin, L.-H. Peng, F.-T. Cheng, and S.-Y. Shang,
  ``Convolutional neural networks for automatic virtual metrology,'' \emph{IEEE
  Robot. Autom. Lett.}, vol.~6, no.~3, pp.~5720--5727, Jul. 2021.

\bibitem{liu2024hierarchical}
Y.-J. Liu, D.~Ni, X.~Shao, D.-L. Gong, and J.-J. Li, ``A hierarchical
  model-based method for wafer level virtual metrology under process
  information deficiency,'' \emph{Qual. Eng.}, vol.~36, no.~3, pp.~534--547,
  Jul. 2024.

\bibitem{hsieh2022transfer}
Y.-M. Hsieh, T.-J. Wang, C.-Y. Lin, Y.-F. Tsai, and F.-T. Cheng,
  ``Convolutional autoencoder and transfer learning for automatic virtual
  metrology,'' \emph{IEEE Robot. Autom. Lett.}, vol.~7, no.~3, pp.~8423--8430,
  Jul. 2022.

\bibitem{hollmann2023tabpfn}
\BIBentryALTinterwordspacing
N.~Hollmann, S.~M{\"u}ller, K.~Eggensperger, and F.~Hutter, ``{TabPFN}: A
  transformer that solves small tabular classification problems in a second,''
  in \emph{Proc. Int. Conf. Learn. Represent. ({ICLR})}, Kigali, Rwanda, May 2023,
  pp.~2859--2894. [Online].
  Available: \url{https://iclr.cc/virtual/2023/poster/12113}
\BIBentrySTDinterwordspacing

\bibitem{hollmann2025foundation}
N.~Hollmann et~al., ``Accurate predictions on small data with a tabular
  foundation model,'' \emph{Nature}, vol.~637, no.~8045, pp.~319--326, Jan. 2025.

\bibitem{zavattari2026warpage}
C.~Zavattari, D.~Demaria, F.~Bonda, and G.~Iacca, ``A machine learning pipeline
  for silicon wafer warpage prediction in diamond coated wire slicing,''
  \emph{Mater. Today Commun.}, vol.~51, Feb. 2026, {Art. no.}~114857.

\bibitem{chen2026physicscmp}
W.~Chen et~al., ``Physics-guided feature construction for small-sample
  virtual metrology in chemical mechanical polishing,'' \emph{IEEE Trans.
  Semicond. Manuf.}, early access, 2026, \doi{10.1109/TSM.2026.3747807}.

\bibitem{jager2026tabpfn35}
\BIBentryALTinterwordspacing
B.~J{\"a}ger et~al., ``{TabPFN-3.5}: Technical report,''
  2026, {arXiv}:2609.17895v3. [Online]. Available:
  \url{https://arxiv.org/abs/2609.17895v3}
\BIBentrySTDinterwordspacing

\bibitem{choi2021film}
J.~E. Choi and S.~J. Hong, ``Machine learning-based virtual metrology on film
  thickness in amorphous carbon layer deposition process,'' \emph{Meas.:
  Sens.}, vol.~16, Aug. 2021, {Art. no.}~100046.

\bibitem{deivendran2025cmp}
B.~Deivendran, V.~S. Masampally, N.~R.~V. Nadimpalli, and V.~Runkana, ``Virtual
  metrology for chemical mechanical planarization of semiconductor wafers,''
  \emph{J. Intell. Manuf.}, vol.~36, no.~3, pp.~1923--1942, Mar. 2025.

\bibitem{chien2022decision}
C.-F. Chien, W.-T. Hung, C.-W. Pan, and T.~H.~V. Nguyen, ``Decision-based
  virtual metrology for advanced process control to empower smart production
  and an empirical study for semiconductor manufacturing,'' \emph{Comput. Ind.
  Eng.}, vol.~169, Jul. 2022, {Art. no.}~108245.

\bibitem{ji2025alignment}
Y.~G. Ji, I.~H. Kim, M.~Shin, G.~T. Lee, and J.~Jang, ``Advanced virtual
  metrology using distribution alignment loss for practical quality control,''
  \emph{Int. J. Prod. Res.}, vol.~63, no.~23, pp.~9051--9069, Dec. 2025.

\bibitem{gumbarevic2026heat}
S.~Gumbarevic, ``Synthetic-to-real transfer with a causal temporal
  convolutional network for site-calibrated virtual heat-flux sensing,''
  \emph{Eng. Proc.}, vol.~163, no.~1, Oct. 2026, {Art. no.}~42.

\bibitem{zhao2025virtual}
M.~Zhao, C.~Taal, S.~Baggerohr, and O.~Fink, ``Graph neural networks for
  virtual sensing in complex systems: Addressing heterogeneous temporal
  dynamics,'' \emph{Mech. Syst. Signal Process.}, vol.~230, May 2025, {Art. no.}~112544.

\bibitem{hsieh2024samples}
Y.-M. Hsieh et~al., ``Developing the keep-important-samples scheme for
  training the advanced {CNN}-based automatic virtual metrology models,''
  \emph{IEEE Robot. Autom. Lett.}, vol.~9, no.~9, pp.~7931--7938, Sep. 2024.

\bibitem{lin2025fewshot}
C.-Y. Lin, T.-L. Tseng, S.~H. Emon, and T.-H. Tsai, ``Large pre-trained models
  and few-shot fine-tuning for virtual metrology: A framework for
  uncertainty-driven adaptive process control in semiconductor manufacturing,''
  \emph{IEEE Trans. Autom. Sci. Eng.}, vol.~22, pp.~15\,351--15\,370, 2025.

\bibitem{muller2022bayesian}
\BIBentryALTinterwordspacing
S.~M{\"u}ller, N.~Hollmann, S.~{Pineda Arango}, J.~Grabocka, and F.~Hutter,
  ``Transformers can do {Bayesian} inference,'' in \emph{Proc. Int. Conf.
  Learn. Represent. ({ICLR})}, virtual conference, Apr. 2022,
  pp.~589--611. [Online]. Available:
  \url{https://iclr.cc/virtual/2022/poster/6595}
\BIBentrySTDinterwordspacing

\bibitem{nikzad2026frp}
M.~H. Nikzad and A.~Ansaripour, ``A dynamic mixture-of-experts framework for
  predicting tensile strength retention of {FRP} composites at elevated
  temperatures,'' \emph{Mater. Des.}, vol.~272, Dec. 2026, {Art. no.}~117337.

\bibitem{liu2026concrete}
H.~Liu, Y.~Wei, J.~Xiao, H.~Yang, W.~Chen, and M.~Bai, ``A prediction method
  for the non-steady-state chloride migration coefficient of {3D}-printed
  concrete integrating {TVAE}-based data augmentation and {PCA}-based
  dimensionality reduction,'' \emph{J. Build. Eng.}, vol.~132, Sep. 2026, {Art. no.}~117475.

\bibitem{grinsztajn2025tabpfn25}
\BIBentryALTinterwordspacing
L.~Grinsztajn et~al., ``{TabPFN-2.5}: Advancing the state of the art in
  tabular foundation models,'' 2025, {arXiv}:2511.08667v2. [Online]. Available: \url{https://arxiv.org/abs/2511.08667v2}
\BIBentrySTDinterwordspacing

\bibitem{grinsztajn2026tabpfn3}
\BIBentryALTinterwordspacing
L.~Grinsztajn et~al., ``{TabPFN-3}: Technical report,''
  2026, {arXiv}:2605.13986v3. [Online]. Available:
  \url{https://arxiv.org/abs/2605.13986v3}
\BIBentrySTDinterwordspacing

\end{thebibliography}
